\documentclass[letterpaper]{article} 
\usepackage[submission]{aaai23}  
\usepackage{times}  
\usepackage{helvet}  
\usepackage{courier}  
\usepackage[hyphens]{url}  
\usepackage{graphicx} 
\usepackage{natbib}  
\usepackage{caption} 
\usepackage{algorithm}
\usepackage{algorithmic}
\usepackage{multirow}
\usepackage{newfloat}
\usepackage{listings}
\DeclareCaptionStyle{ruled}{labelfont=normalfont,labelsep=colon,strut=off} 
\floatstyle{ruled}
\newfloat{listing}{tb}{lst}{}
\floatname{listing}{Listing}
\title{Neuro-symbolic Meta Reinforcement Learning for Trading}
\author{
    S I Harini,\textsuperscript{\rm 1}
    Gautam Shroff,\textsuperscript{\rm 2}
    Ashwin Srinivasan,\textsuperscript{\rm 1}
    Prayushi Faldu,\textsuperscript{\rm 3}
    Lovekesh Vig \textsuperscript{\rm 2}
}
\affiliations{
    \textsuperscript{\rm 1} APPCAIR, BITS Pilani K.K. Birla Goa Campus\\
    \textsuperscript{\rm 2} TCS Research\\
    \textsuperscript{\rm 3} IIT Delhi\\
}

\begin{document}

\maketitle

\begin{abstract}

We model short-duration (e.g. day) trading in financial
markets as a sequential decision-making problem under uncertainty,
with the added complication of continual concept-drift.
We, therefore, employ meta reinforcement learning via the RL$^2$ algorithm.
It is also known that human traders often rely on frequently occurring 
symbolic patterns in price series. We employ logical program induction
to discover symbolic patterns that occur frequently as well as recently,
and explore whether using such features improves the performance
of our meta reinforcement learning algorithm. We report experiments
on real data indicating that meta-RL is better than vanilla RL and
also benefits from learned symbolic features.
\end{abstract}

\section{Introduction}
Deep learning techniques have achieved human-like and even superhuman performance in a number of arenas: video games, 
strategy games, robotics, etc. In many of these arenas, the spectrum
of human performance varies widely, from average to expert. Human traders in financial markets also differ greatly
in skill and performance. The consistent success of expert traders is unlikely to be due to chance alone; it 
is more likely that such traders are explicitly or implicitly relying on patterns in the data they see. 

External events in the world clearly affect prices in financial markets. Thus, accurately forecasting financial prices over medium
term, e.g., months, weeks or even days is a challenging if not hopeless task, even for deep-learning techniques, see \cite{shah2021forecasting}.
The impact of external events is reduced when trading over a shorter time frame, such as sub-second high-frequency trading
as well as, to an extent, intra-day trading. However, intra-day price variations are the result of complex feedback
between buyers and sellers, as modeled in \cite{wang2014dynamical}.
Such feedback results in near chaotic, but not random, behavior, i.e.,
the auto-correlation in the price series is not zero: the price series has `memory'. 
Thus, short-duration price series, while near chaotic, do seem to exhibit patterns albeit not consistently, which is what human traders exploit.

It is known that deep-learning techniques can indeed track chaotic systems, such as the Lorenz equations \cite{madondo2018learning}; 
thus, it is not unreasonable to expect deep learning could be effective in near chaotic environments such as financial markets.
Indeed, there have been many attempts at applying machine learning, and, more recently, deep learning to trading, some of which
are mentioned in the next section.

Even so, applying deep learning in financial markets faces challenges: First, data
is scarcer than one may expect - there are only so many days of past history, and so many financial instruments, even if one
assumes all instruments behave alike. This is far fewer data than, say, text or images on the internet. Second, price series exhibit
patterns, these change over time, and it is well known that markets continually change their `regime', e.g. from `trending' to `mean-reverting',
as well as their volatility. 

In this paper we (a) model trading as one of sequential decision-making under uncertainty, (b) apply {\bf deep meta reinforcement learning} to
make trading decisions and (c) investigate whether the incorporation of features based on {\bf hand-crafted patterns} that are often used
by human traders improves performance. Further, we observe a {\bf meta-pattern} in such hand-crafted patterns which we use to
automatically learn a large number of similar features using techniques borrowed from {\bf inductive logic programming}, and investigate
whether these add to the effectiveness of our meta-RL based trading agent. We present \emph{preliminary} results on real data that indicate that both meta reinforcement learning and logical features, both hand-crafted and learned, are more effective than vanilla RL or primary price features alone. We conclude with ideas for future  exploration.

\section{Background/Related Work}

\subsection{Meta-learning}
Meta-learning approaches seek to learn in situations training data is scarce,
either inherently or due to rapid distribution shifts that render older data less relevant to the current task,
as is the case for short-duration trading.
Meta-learning techniques are` learning to learn' by training on many related tasks, so that
performance on similar future tasks is improved. Optimization-based meta-learning, exemplified
by the MAML algorithm \cite{finn2017model} attempts to learn a good parameter initialization such that
a few steps of gradient descent starting from there are sufficient to adapt rapidly to a new task even
with very few data samples. While MAML and related meta-learning techniques do apply in reinforcement learning
setting, they still require training on a new task, albeit with limited data. In the case of trading, where
an episode is ideally an entire day, it is of limited use since a model so adapted could only be used the next day,
and without further adaptation as the day progresses. (Metric-based meta-learning techniques, such as matching networks
\cite{vinyals2016matching} 
do not easily apply in reinforcement learning.) The third class of traditional
meta-learning techniques are model-based, wherein adaption on new data takes place within the activations of a
network's hidden states rather than via any gradient-based updates. The RL$^2$ algorithm \cite{duan2016rl} which we use here is also a 
model-based meta-learning technique. In such a technique, adaption on new data, in our case new rewards, takes place
within the network activations as rewards arrive, making such a technique most applicable in our scenario.
\subsection{Machine-learning for Trading}
Recent works applying machine learning to trading based on price signals alone have also used deep neural networks
as well as reinforcement learning: `Deep Momentum Networks' \cite{lim2019enhancing} as well as `Momentum Transformer'
\cite{wood2021trading} formulate the trading task as one of suggesting the position to take, e.g., 1 for a long
(i.e., buy) position, -1 for a short (i.e., sell) position, and 0 for no position. Exiting a buy/sell position 
takes place when a 0 action follows the previous 1 actions, etc. Neural networks are trained to directly
optimize volatility-adjusted expected returns, adjusted for transaction costs, over a trading period. 
Transaction costs are computed by tracking when positions are entered and exited, i.e., when actions change
from one time step to the next. The former paper uses MLPs and LSTMs, while the latter uses transformers.
Both works are essentially applying vanilla REINFORCE to the MDP formulation of the trading problem.
Deep Reinforcement Learning in Trading \cite{zhang2020deep} uses the same formulation as the above two works,
but applies more refined reinforcement learning techniques, e.g., policy-gradient, actor-critic, and deep-Q-learning algorithms.
In contrast to the formulation used in all the above three works, we also model actions as buy (1), sell(-1), or do nothing (0),
but these can \emph{only} be taken when no position (i.e., zero shares) are held. As soon as a position is taken
our training environment computer when this position exits due to the pre-defined stop-loss/target being met or the day end
being reached, at which point a reward is returned to the RL agent. We postulate that such a formulation
makes for easier learning since the agent only needs to deal with one kind of situation, i.e. when it holds no
position. The downside is that exit conditions (i.e., stop-loss/targets/end-of-day) are fixed in advance, rather
than determined based on price movements. Note that however, such exit conditions can also be outputs of the policy, 
i.e., varying stop-loss/targets for each buy or sell, based on current volatility, or whatever the agent finds useful;
nevertheless, we have not reported experiments with this enhancement here.
\subsection{Inductive Logic Programming}
Inductive logic programming (ILP) \cite{Muggleton1994InductiveLP} investigates the inductive construction of first-order 
clausal theories from examples and background knowledge. ILP is ordinarily employed in a supervised learning setting, where positive (and usually
negative) examples of a target concept are given in terms of base features. Also supplied is background knowledge in the form of
facts as well as logical rules, typically in a logic programming language, i.e., Prolog. The ILP process involves
constructing a theory that explains the examples provided with the desired accuracy, support, and confidence. 
At each stage in this process, possible theories are tested against target examples via resolution using background knowledge.

In our case we do not have target examples or concepts, instead, price data is translated into Prolog facts, and
feature \emph{templates}, or `meta-rules', are added as background knowledge using techniques introduced in \cite{muggleton2015meta}. 
Thereafter, starting from instances selected randomly as in \cite{Muggleton1994InductiveLP}, features
with high support in the data are discovered as in \cite{dehaspe1997mining}, via (SLD) resolution\footnote{https://en.wikipedia.org/wiki/SLD\_resolution} 
using the supplied background facts and meta-rules. 

\section{Methodology}
\subsection{Task Formulation and Learning Environment}
Each task represents a trading day for a particular symbol (i.e., stock). Data arrives each minute with the open, high, low, and close prices
for the past minute along with technical analysis indicators (as will be detailed below in a subsequent section). 
The agent issues buy/sell/do-nothing actions based on the data seen so far for the day. Once an order (i.e., buy or sell) is placed,
the agent does not see any data (or reward) until the price moves by an amount determined by pre-defined stop-loss or target values; e.g. if these 
are each 1\%, the agent receives a reward (positive or negative) when the close price changes by 1\% from the point at which the order was placed,
At this point, the agent resumes receiving data every minute until it places another order. Alternatively, if the day ends, the agents receive a reward
based on the final closing price of the day. As noted earlier, this formulation differs from that in prior works \cite{lim2019enhancing,wood2021trading,zhang2020deep}.
\subsection{Meta-reinforcement Learning: RL$^2$}
To deal with continual distribution shift, we employ the meta reinforcement learning approach RL$^2$ from \cite{duan2016rl}. In a standard RL formulation
the agent predicts the next action based on the current state (or history of states, in the case of a recurrent network) and subsequently receives a reward.
In RL$^2$, the previous action and reward are also input to the network, and a recurrent network is used. As a result, changes in the \{state,action,reward\}
distribution are visible to the agent as it takes actions. Note that  the agent is trained over many trials where it encounters sequences of tasks with possibly different \{state, action, reward\} distributions. Thus, in principle, the agent can learn to adapt to a new distribution when encountering a new task.

The RL$^2$ agent is trained on past data comprising of trials, where each trial is a sequence of tasks. In the trading context, this entails training 
the agent over many day-symbol combinations, and then testing for a new day (for one of the symbols already seen; though unseen symbols could also be used - 
here we use seen symbols). In principle, the meta-RL agent should rapidly adapt to the reward pattern it experiences in its first few orders even if these
differ from the recent past. We use PPO to train the meta-RL agent and an LSTM agent (whereas \cite{duan2016rl} used TRPO and a GRU agent), since
PPO is known to be more stable while training and LSTMs are more expressive than GRUs.

\subsection{Hand-crafted Features}
\begin{figure}[h]
    \centering
    \includegraphics[scale=0.4]{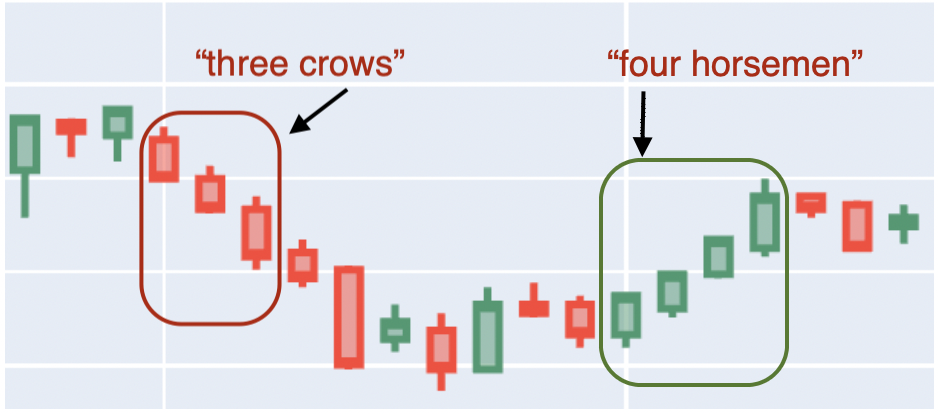}
    \caption{Example of hand-crafted features}
    \label{fig:handcrafted_features}
\end{figure}

Human traders use price patterns as signals on which to base their trading decisions. An example of two such patterns
are depicted in Figure \ref{fig:handcrafted_features}. The `three crows' pattern involves three successive observations 
over which both opening prices and closing prices decrease sequentially. Similarly, the 'four horsemen' pattern involves
a sequence of four rising open/close prices. 
In order to explore whether such features add additional value we detect the presence or absence of such patterns
and append two Boolean features to the state at each time step to indicate if and which of these holds: thus
these features would be $(0,0)$ for all time steps except at time step 17 where a $(0,1)$ would indicate the presence
of `four horsemen', and at 6,7, and 8 where $(1,0)$ would indicate the presence of `three crows' over three recent steps.
\subsection{Learned Logical Features}
The two handcrafted features above are based on patterns involving increasing/decreasing sequences of primary features, viz., open and close prices
respectively. We postulate that increasing and decreasing sequences of other primary features might also form useful features. For example,
increasing/decreasing sequences of highs, lows, or even moving averages or other technical indicators. 

Further, it is also possible that
increasing/decreasing sequences of \emph{derived} features, in particular, differences between primary features may be useful. For example,
whether the difference between open and close is narrowing or widening may be indicative of decreasing or increasing volatility, which in turn
should be useful for deciding an appropriate action. The same can be said for differences in highs and lows, opens and highs, etc. It 
is also reasonable to consider differences between technical indicators as well, e.g., differences between moving averages of different lengths;
the crossing of such moving averages is known to be used by traders, so narrowing differences would point to a possible impending crossing. 

Of course, exhaustively enumerating all such differences for each time step would be both computationally inefficient and likely to
confuse the neural network model. Instead, we use techniques borrowed from inductive logic programming to enumerate such features in a principled
manner and filter these first based on the frequency of occurrence in the training data, and then by importance using standard feature-importance
determinants.

{\bf Meta-patterns} are defined in Prolog to capture the concept of `runs', i.e., a sequence of continually increasing/decreasing values
(of one or more primary/derived features) of arbitrary length. Thereafter randomized search followed by resolution in Prolog is used to enumerate frequently
occurring patterns (i.e., those that occur in the training data more often than a given support value). Background clauses are included
to define the concept of a `derived' feature as the difference between two primary features. Randomness is used in the search process to
select features to test for. Search proceeds until the number of high support patterns found reaches a pre-determined maximum limit.

The above procedure yields a very large number of patterns, which are used to augment the price data with pattern-based features determined by the
presence or absence of one or more patterns at each time step. These features are used to train a random forest regression model to 
predict reward; as a side-effect, the random forest model returns the importance of each feature towards predicting reward. These
importance values are used to rank pattern-based features. Finally, a small number of top-ranking features are used to augment the meta-RL 
agent's neural network model.

\section{Evaluation}
\subsection{Data}
Data received by the agent at each time step is price data, i.e., open, high, low, close, and volume. These are normalized
by dividing the volume column by the first non zero volume at the beginning of the episode and the rest by dividing by the close price at beginning of the episode.
These normalized values are then used to compute technical analysis indicators such as simple moving averages, relative strength indicator,s etc. We use
a total of 15 such technical analysis features in addition to open, high, low, close prices and volume.
\subsection{Results}
The agent is trained on $n$ symbols for $m$ days and test scores on the day $m+1$ are calculated. The results averaged over multiple non-overlapping subsets of symbols and days are as shown in Table \ref{tab:rlvsmetarl} and Table \ref{tab:featureperformance}. Each table entry indicates the \% average daily return achieved on
the test day when trained using the data of the given number of symbols and previous days. Table \ref{tab:rlvsmetarl} shows the performance of vanilla reinforcement
learning (i.e., wherein the previous action and reward is \emph{not} fed back into the neural network, vs meta reinforcement learning, and Table \ref{tab:featureperformance} shows the performance when using different sets of features as inputs
to the RL-agent's neural network.

\begin{table}[h]
\begin{tabular}{|l|l|l|l|l|}
\hline
                            &         & 1 symbol & 3 symbols & 6 symbols \\ \hline
\multirow{3}{*}{Meta RL}    & 5 days  & 0.26     & -0.23     & 0.36      \\ \cline{2-5} 
                            & 10 days & -0.06    & 0.11      & 0.27      \\ \cline{2-5} 
                            & 15 days & 0.11     & 0.04      & 0.09      \\ \hline
\multirow{3}{*}{Vanilla RL} & 5 days  & -0.12    & -0.33     & -0.45     \\ \cline{2-5} 
                            & 10 days & -0.17    & -0.31     & -0.52     \\ \cline{2-5} 
                            & 15 days & -0.27    & -0.22     & -0.38     \\ \hline
\end{tabular}
\caption{Avg. \% returns on test day using vanilla/meta RL.}
\label{tab:rlvsmetarl}
\end{table} 
\begin{table}[h]
\centering
\begin{tabular}{|l|l|l|l|l|}
\hline
                                      &         & 1 symbol & 3 symbols & 6 symbols \\ \hline
\multirow{3}{40 pt}{Technical Indicator \\features}              & 5 days  & 0.26     & -0.23     & 0.36      \\ \cline{2-5} 
                                      & 10 days & -0.06    & 0.11      & 0.27      \\ \cline{2-5} 
                                      & 15 days & 0.11     & 0.04      & 0.09      \\ \hline
\multirow{3}{50 pt}{+Handcrafted features} & 5 days  & 0.32     & -0.16     & 0.41         \\ \cline{2-5} 
                                      & 10 days & 0.07     & 0.21      & 0.19         \\ \cline{2-5} 
                                      & 15 days & 0.15     & 0.12      & 0.14         \\ \hline
\multirow{3}{50 pt}{+Learned logical\\features}     & 5 days  & 0.31     & 0.27      & 0.45      \\ \cline{2-5} 
                                      & 10 days & 0.17     & 0.39      & 0.21      \\ \cline{2-5} 
                                      & 15 days & -0.1     & 0.23      & 0.11      \\ \hline
\multirow{3}{50 pt}{+Handcrafted\\+logical features}  & 5 days  & 0.33        & 0.36         & 0.29         \\ \cline{2-5} 
                                      & 10 days & 0.19        & 0.29         & 0.25         \\ \cline{2-5} 
                                      & 15 days & -0.04        & 0.27         & 0.05         \\ \hline
\end{tabular}
\caption{Avg. \% returns on test day for different feature sets. }
\label{tab:featureperformance}
\end{table}

\section{Discussion}
We draw the following indicative conclusions from the results presented above:
\begin{enumerate}
    \item Meta reinforcement learning improves over vanilla reinforcement learning, indicating that distribution shift may
    be impacting the former.
    \item Logical features, both hand-crafted as well as learned, improve performance vs using primary features alone.
    \item {\bf Learned logical features add value over and above hand-crafted features alone.}
    \item Training using too much past data (15 days) is inferior to training using a moderate
    amount of past data (5 or 10 days). This may be further evidence of distribution shift over longer time periods.
    \item Training on more symbols is better, indicating that distribution shifts take place more over days rather than across different symbols.
\end{enumerate}

\section{Conclusions and Future Work}
We submit that meta reinforcement learning is a promising direction to explore for building trading agents using deep learning.
Also, logical features learned using meta-patterns inspired by hand-crafted features may be useful. 

Many recent advances in deep learning are worthwhile exploring in the context of building trading agents: Language models have proven
to be few-shot learners even for numerical data expressed symbolically \cite{hegselmann2022tabllm}; could logical features as we have used here form the basis 
for exploring whether these systems could be applied in the trading arena? 
Recently de-noising diffusion models have also been used for planning \cite{janner2022planning}; trying such approaches in trading may be worth exploring as well.
\bibliography{refs}

\end{document}